\documentclass{article}
\usepackage[utf8]{inputenc}
\usepackage{pdflscape}
\usepackage{epstopdf} 
\usepackage[hyphens]{url}
\usepackage{todonotes}
\usepackage{subfigure}
\usepackage[english]{babel}
\usepackage[export]{adjustbox}
\RequirePackage{atbegshi}
\usepackage{floatrow}
\usepackage[T1]{fontenc}
\usepackage{lmodern}
\usepackage{setspace}
\usepackage{listings}
\usepackage{microtype}
\usepackage[T1]{fontenc}
\usepackage{amsmath,mathtools}
\usepackage{rotating} 
\graphicspath{{}}
\usepackage{color}
\usepackage{graphicx}
\usepackage{framed}
\ProvideTextCommand{\textasciitilde}{OT1}{\~{}}

\title{Binary Particle Swarm Optimization versus Hybrid Genetic Algorithm for Inferring Well Supported Phylogenetic Trees}

\author{Bassam~AlKindy, Bashar~Al-Nuaimi, Christophe~Guyeux, Jean-Fran\c{c}ois\\Couchot, Michel~Salomon, Reem~Alsrraj and Laurent~Philippe}

\begin{document}
\maketitle
\setcounter{footnote}{0}

\begin{abstract}
The amount of completely sequenced chloroplast genomes increases rapidly every day, leading to the possibility to build large-scale phylogenetic trees of plant species. Considering a subset of close plant species defined according to their chloroplasts, the phylogenetic tree that can be inferred by their core genes is not necessarily well supported, due to the possible occurrence of “problematic” genes (i.e., homoplasy, incomplete lineage sorting, horizontal gene transfers, etc.) which may blur the phylogenetic signal. 
However, a trustworthy phylogenetic tree can still be obtained 
provided such a number of blurring genes is reduced.
The problem is thus  to determine the largest subset of core genes that produces the best-supported tree.
To discard problematic genes and due to the overwhelming number of possible combinations, this article focuses on how to extract the largest subset of sequences in order to obtain the most supported species tree. Due to computational complexity, a distributed Binary Particle Swarm Optimization (BPSO) is proposed in sequential and distributed fashions. Obtained results from both versions of the BPSO are compared with those computed using an hybrid approach embedding both genetic algorithms and statistical tests. 
The proposal has been applied to different cases of plant families, leading to encouraging results for these families.
\end{abstract}

\section{Introduction}\label{sec:intro}

The multiplication of completely sequenced chloroplast genomes should normally lead to the ability to infer reliable phylogenetic trees for plant species. This is due to the existence of trustworthy coding sequence prediction and annotation software specific to chloroplasts (like DOGMA~\cite{RDogma}) and of accurate sequence alignment tools.
Additionally, given a set of biomolecular sequences or characters, various well-established approaches have been developed in recent years to deduce their phylogenetic relationship, encompassing methods based on Bayesian inference or maximum likelihood~\cite{Stamatakis21012014}. 
Robustness aspects of the produced trees can be evaluated too, for instance through bootstrap analyses. 
In other words, given a set of close plant species, their core genome (the set of genes in common) is as large and accurately detected as possible, to hope to be able to finally obtain a well-supported phylogenetic tree. However, all genes of the core genome are not necessarily constrained in a similar way, some genes having a larger ability to evolve than other ones due to their lower importance: such minority genes tell their own story instead of the species one, blurring so the phylogenetic information. 
The link between the robustness and accuracy of the phylogenetic tree, and the amount of data used for this reconstruction, is not yet completely understood. More precisely, if we consider a set of species reduced to lists of gene sequences, we have an obvious dependence between the chosen subset of sequences and the obtained tree (topology, branch length, and/or robustness). 
This dependence is usually regarded by the mean of gene trees merged in a phylogenetic network.
This article investigates the converse approach: it starts by the union of whole core genes and tries to remove the ones that blear the phylogenetic signals. More precisely, the objective is to find the largest part of the genomes that produces a phylogenetic tree as supported as possible, reflecting by doing so the relationship of the largest part of the sequences under consideration. 

Due to an overwhelming number of combinations to investigate, a brute force approach is a nonsense, which explains why heuristics are considered. 

A previous work~\cite{genetic2015} has proposed the use of an ad hoc Genetic Algorithm (GA) to solve the problem of finding the largest subset of core genes producing a phylogenetic tree as supported as possible. However, in some situations, this algorithm fails to solve the optimization problem due to a low convergence rate. The proposal of this research work is thus to investigate the application of the Binary Particle Swarm Optimization (BPSO) to face our optimization challenge, and to compare it to the GA one. A new algorithm has been proposed and applied, in a distributed manner using supercomputing facilities, to investigate the phylogeny of various families of plant species.

This article is indeed an extended and improved version of the work published in the CIBB proceedings book~\cite{aagp+15:ip}. New contributions encompass a second version of the BPSO for phylogenetic studies together with its distributed algorithm. The two BPSO versions are evaluated on a large number of new group of species.
New experimental results have been thus obtained
with this BPSO based approaches and with the genetic algorithm and 
further compared. 

The remainder of this article is organized as follows.
Section~\ref{Problem review} gives a general presentation of the problem, 
further recalls how to extract the restricted set of core genes, and
next presents various tools for constructing the phylogenetic tree from the hybrid approach. It ends with a brief description of the BPSO metaheuristic. 
Section~\ref{sec:PSPI} describes the way 
the metaheuristic approach is applied to solve problematic supports in biomolecular based phylogenies, considering the particular case of \emph{Rosales} order. The distributed version of BPSO algorithm using MPI is also discussed. 
Obtained results and comparisons with GA approaches are detailed in Section~\ref{sec:exp}. Finally, this paper ends with a conclusion section, in which the article is summarized and intended future work is outlined.

\section{Presentation of the problem}\label{Problem review}

\subsection{General presentation}
Let us consider a set of chloroplast genomes that have been annotated using DOGMA~\cite{RDogma}.
Following~\cite{Alkindy2014,Alkindy_BIBM2014}, we have then access to the restricted core genome~\cite{Alkindy2014} (genes present everywhere) of these species, whose size is about one hundred genes when the species are close enough.
Sequences are further aligned using MUSCLE~\cite{edgar2004muscle} and the RAxML~\cite{Stamatakis21012014} tool infers the corresponding phylogenetic tree. If the resulting tree is well-supported (\textit{i.e.}, if all bootstrap values are larger than 95) we
can indeed reasonably consider that the phylogeny of these species is resolved.

\begin{figure}[tb]
\begin{center}
    \includegraphics[scale=0.5]{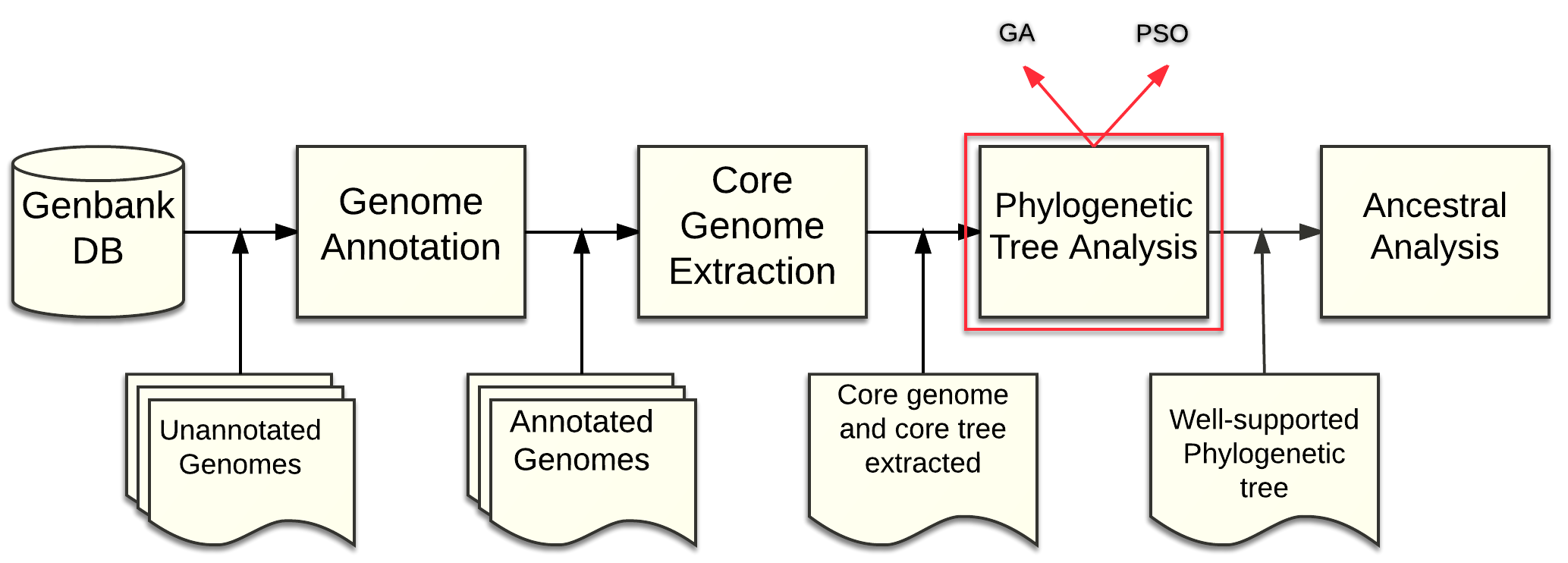}
    \caption{Overview of the proposed pipeline}\label{fig:GA_generalview}
\end{center}
\end{figure}

In a case where some branches are not well supported, we can wonder whether a few genes can be incriminated in this lack of support. If so, we face an optimization problem: \emph{find the most supported tree using the largest subset of core genes}. Obviously, a brute force approach investigating all possible combinations of genes is intractable, as it leads to $2^n$ phylogenetic tree inferences for a core genome of size $n$. 
To solve this optimization problem, we have formerly proposed in~\cite{genetic2015} a general pipeline detailed in Figure~\ref{fig:GA_generalview}.
In this pipeline, the stage of phylogenetic tree analysis mixes both genetic algorithm with LASSO tests in order to discover problematic genes. 
However, deeper experimental investigations summarized in Table~\ref{tab:families}
have shown that such a pipeline does not succeed to predict the phylogeny of some particular plant orders: in 14 groups of species the pipeline produces a score 
of bootstrap lower than 95 (the $b$ column). It is important to understand what the bootstrap value represents before we can get a good response for what is "good" or "poor" support. 

Bootstrapping is a resembling analysis that involves taking columns of characters out of the analysis, rebuilding the tree, and testing if the same nodes are recovered. This is done through many (100 or 1000, quite often) iterations. If, for example, you recover the same node through 95 of 100 iterations of taking out one character and resampling your tree, then you have a good idea that the node is well supported. If we get low support, this suggests that only few characters support that node, as removing characters at random from your matrix leads to a different reconstruction of that node. 

We thus wonder whether a binary particle swarm optimization approach can outperform the GA when finding the largest subset of core genes producing the most supported phylogenetic tree (GA replaced by the BPSO in the ``Phylogenetic tree analysis'' box of Figure~\ref{fig:GA_generalview}). 

Let us now give the general idea behind particle swarm optimization.
\begin{table}[tb]
\tiny
\centering
\caption{Results of genetic algorithm approach on various families.}\label{tab:families}
\begin{tabular}{l||c|c|c|c|c|c|l}
\multicolumn{1}{c||}{Group} & occ & $c$ & \# taxa & $b$ & Terminus & Likelihood & \multicolumn{1}{c}{Outgroup} \\ \hline\hline
\textit{Gossypium\_group\_0} & 85 & 84 & 12 & 26 & 1 & -84187.03 & \textit{Theo\_cacao} \\ 
\textit{Ericales} & 674 & 84 & 9 & 67 & 3 & -86819.86 & \textit{Dauc\_carota} \\ 
\textit{Eucalyptus\_group\_1} & 83 & 82 & 12 & 48 & 1 & -62898.18 & \textit{Cory\_gummifera} \\ 
\textit{Caryophyllales} & 75 & 74 & 10 & 52 & 1 & -145296.95 & \textit{Goss\_capitis-viridis} \\ 
\textit{Brassicaceae\_group\_0} & 78 & 77 & 13 & 64 & 1 & -101056.76 & \textit{Cari\_papaya} \\ 
\textit{Orobanchaceae} & 26 & 25 & 7 & 69 & 1 & -19365.69 & \textit{Olea\_maroccana} \\ 
\textit{Eucalyptus\_group\_2} & 87 & 86 & 11 & 71 & 1 & -72840.23 & \textit{Stoc\_quadrifida} \\ 
\textit{Malpighiales} & 422 & 78 & 10 & 96 & 3 & -91014.86 & \textit{Mill\_pinnata} \\ 
\textit{Pinaceae\_group\_0} & 76 & 75 & 6 & 80 & 1 & -76813.22 & \textit{Juni\_virginiana} \\ 
\textit{Pinus} & 80 & 79 & 11 & 80 & 1 & -69688.94 & \textit{Pice\_sitchensis} \\ 
\textit{Bambusoideae} & 83 & 81 & 11 & 80 & 3 & -60431.89 & \textit{Oryz\_nivara} \\ 
\textit{Chlorophyta\_group\_0} & 231 & 24 & 8 & 81 & 3 & -22983.83 & \textit{Olea\_europaea} \\ 
\textit{Marchantiophyta} & 65 & 64 & 5 & 82 & 1 & -117881.12 & \textit{Pice\_abies} \\ 
\textit{Lamiales\_group\_0} & 78 & 77 & 8 & 83 & 1 & -109528.47 & \textit{Caps\_annuum} \\ 
\textit{Rosales} & 81 & 80 & 10 & 88 & 1 & -108449.4 & \textit{Glyc\_soja} \\ 
\textit{Eucalyptus\_group\_0} & 2254 & 85 & 11 & 90 & 3 & -57607.06 & \textit{Allo\_ternata} \\ 
\textit{Prasinophyceae} & 39 & 43 & 4 & 97 & 1 & -66458.26 & \textit{Oltm\_viridis} \\ 
\textit{Asparagales} & 32 & 73 & 11 & 98 & 1 & -88067.37 & \textit{Acor\_americanus} \\ 
\textit{Magnoliidae\_group\_0} & 326 & 79 & 4 & 98 & 3 & -85319.31 & \textit{Sacc\_SP80-3280} \\ 
\textit{Gossypium\_group\_1} & 66 & 83 & 11 & 98 & 1 & -81027.85 & \textit{Theo\_cacao} \\ 
\textit{Triticeae} & 40 & 80 & 10 & 98 & 1 & -72822.71 & \textit{Loli\_perenne} \\ 
\textit{Corymbia} & 90 & 85 & 5 & 98 & 2 & -65712.51 & \textit{Euca\_salmonophloia} \\ 
\textit{Moniliformopses} & 60 & 59 & 13 & 100 & 1 & -187044.23 & \textit{Prax\_clematidea} \\ 
\textit{Magnoliophyta\_group\_0} & 31 & 81 & 7 & 100 & 1 & -136306.99 & \textit{Taxu\_mairei} \\ 
\textit{Liliopsida\_group\_0} & 31 & 73 & 7 & 100 & 1 & -119953.04 & \textit{Drim\_granadensis} \\ 
\textit{basal\_Magnoliophyta} & 31 & 83 & 5 & 100 & 1 & -117094.87 & \textit{Ascl\_nivea} \\ 
\textit{Araucariales} & 31 & 89 & 5 & 100 & 1 & -112285.58 & \textit{Taxu\_mairei} \\ 
\textit{Araceae} & 31 & 75 & 6 & 100 & 1 & -110245.74 & \textit{Arun\_gigantea} \\ 
\textit{Embryophyta\_group\_0} & 31 & 77 & 4 & 100 & 1 & -106803.89 & \textit{Stau\_punctulatum} \\ 
\textit{Cupressales} & 87 & 78 & 11 & 100 & 2 & -101871.03 & \textit{Podo\_totara} \\ 
\textit{Ranunculales} & 31 & 71 & 5 & 100 & 1 & -100882.34 & \textit{Cruc\_wallichii} \\ 
\textit{Saxifragales} & 31 & 84 & 4 & 100 & 1 & -100376.12 & \textit{Aral\_undulata} \\ 
\textit{Spermatophyta\_group\_0} & 31 & 79 & 4 & 100 & 1 & -94718.95 & \textit{Mars\_crenata} \\ 
\textit{Proteales} & 31 & 85 & 4 & 100 & 1 & -92357.77 & \textit{Trig\_doichangensis} \\ 
\textit{Poaceae\_group\_0} & 31 & 74 & 5 & 100 & 1 & -89665.65 & \textit{Typh\_latifolia} \\ 
\textit{Oleaceae} & 36 & 82 & 6 & 100 & 1 & -84357.82 & \textit{Boea\_hygrometrica} \\ 
\textit{Arecaceae} & 31 & 79 & 4 & 100 & 1 & -81649.52 & \textit{Aegi\_geniculata} \\ 
\textit{PACMAD\_clade} & 31 & 79 & 9 & 100 & 1 & -80549.79 & \textit{Bamb\_emeiensis} \\ 
\textit{eudicotyledons\_group\_0} & 31 & 73 & 4 & 100 & 1 & -80237.7 & \textit{Eryc\_pusilla} \\ 
\textit{Poeae} & 31 & 80 & 4 & 100 & 1 & -78164.34 & \textit{Trit\_aestivum} \\ 
\textit{Trebouxiophyceae} & 31 & 41 & 7 & 100 & 1 & -77826.4 & \textit{Ostr\_tauri} \\ 
\textit{Myrtaceae\_group\_0} & 31 & 80 & 5 & 100 & 1 & -76080.59 & \textit{Oeno\_glazioviana} \\ 
\textit{Onagraceae} & 31 & 81 & 5 & 100 & 1 & -75131.08 & \textit{Euca\_cloeziana} \\ 
\textit{Geraniales} & 31 & 33 & 6 & 100 & 1 & -73472.77 & \textit{Ango\_floribunda} \\ 
\textit{Ehrhartoideae} & 31 & 81 & 5 & 100 & 1 & -72192.88 & \textit{Phyl\_henonis} \\ 
\textit{Picea} & 31 & 85 & 4 & 100 & 1 & -68947.4 & \textit{Pinu\_massoniana} \\ 
\textit{Streptophyta\_group\_0} & 31 & 35 & 7 & 100 & 1 & -68373.57 & \textit{Oedo\_cardiacum} \\ 
\textit{Gnetidae} & 31 & 53 & 5 & 100 & 1 & -61403.83 & \textit{Cusc\_exaltata} \\ 
\textit{Euglenozoa} & 29 & 26 & 4 & 100 & 3 & -8889.56 & \textit{Lath\_sativus} \\ \hline\hline
\end{tabular}
\end{table}

\subsection{Binary Particle Swarm Optimization}\label{Sec:PSO}

Particle Swarm Optimization (PSO) is a stochastic optimization technique developed by Eberhart and Kennedy in 1995~\cite{kenndy1995particle}. 
PSOs have been successfully applied on various optimization problems like function optimization, artificial neural network training, and fuzzy system control. In this metaheuristic, particles follow a very simple behavior that is to learn from the success of neighboring individuals. An emergent behavior enables individual swarm members, particles, to take benefit from the discoveries, or from previous experiences, of the other particles that have obtained more accurate solutions. 
In the case of the standard binary PSO model~\cite{intechopen}, the particle position is a vector of $N$ parameters that can be set as ``yes'' or ``no'', ``true'' or ``false'', ``include'' or ``not include'', \emph{etc.} (binary values). A function associates to such kind of vector a score (real number) according to the optimization problem. The objective is then to define a way to move the particles in the $N$ dimensional binary search space so that they produce the optimal binary vector w.r.t. the scoring function.

In more details, each particle $i$ is represented by a binary vector $X_i$ (its position). Its length $N$ corresponds to the dimension of the search space, that is, the number of binary parameters to investigate. A $1$ in coordinate $j$ of this vector means that the associated $j$-th parameter is selected. A swarm of $n$ particles is then a list of $n$ vectors of positions $\left(X_1, X_2, \dots, X_n\right)$ together with their associated velocities $V = (V_1, V_2, ..., V_n)$, which are $N$-dimensional vectors of real numbers between 0 and 1. These latter are initially set randomly. 
At each iteration, a new velocity vector is computed as follows:
\begin{eqnarray}\label{eq:2}
V_i(t+1)= w V_i(t)+\phi_1\left(P_{i}^{best}-X_{i}\right)+\phi_2\left(P_{g}^{best}-X_{i}\right)
\end{eqnarray}
where $w$, $\phi_1$, and $\phi_2$ are weighted parameters setting the level of each three trends for the particle, which are respectively to continue in its adventurous direction, to move in the direction of its own best position $P_{i}^{best}$, or to follow the gregarious instinct to the global best known solution $P_{g}^{best}$.
Both $P_{i}^{best}$ and $P_{g}^{best}$ are computed according to the scoring function.

The new position of the particle is then obtained using the equation below:
\begin{eqnarray}\label{eq:3}
X_{ij}(t+1)= 
\begin{cases}
      1 & \text{if}\ {\tiny r}_{ij}\leq Sig(V_{ij}(t+1)) , \\
      0 & \text{otherwise},
    \end{cases} 
\end{eqnarray}
where $r_{ij}$ is a threshold that depends on both the particle $i$ and the parameter $j$, while the $Sig$ function is the sigmoid one~\cite{intechopen}, that is:
\begin{eqnarray}\label{eq:1}
\textit{Sig}(V_{ij}(t+1))=\frac{1}{1+e^{-V_{ij}(t+1)}}
\end{eqnarray}
Let us now recall how to use a BPSO approach to solve our optimization problem related to phylogeny~\cite{aagp+15:ip}.

\section{Particle Swarm for Phylogenetic Investigations}\label{sec:PSPI}

\subsection{BPSO applied to phylogeny}
\label{sec:algo}


In order to illustrate how to use the BPSO approach, we have considered the \emph{Rosales} order, which has already been analyzed in~\cite{genetic2015} using a hybrid genetic algorithm and Lasso test approach.
The \emph{Rosales} order is constituted by 9~ingroup species and 1~outgroup (\textit{Mollissima}), as described in Table~\ref{tab:species}. They have been annotated using DOGMA and their core genome has been computed according to the method described in~\cite{Alkindy2014,Alkindy_BIBM2014}.
Its size is equal to 82~genes. 
Unfortunately, the phylogeny cannot be resolved directly neither by considering all these core genes 
nor by considering any of the 82 combinations of 81 core genes. 


\begin{table}[tb]
\caption{Genomes information of \textit{Rosales} species under consideration}
\label{tab:species}
\small
\centering
\scalebox{0.8}{%
\begin{tabular}{l l c l l}
\hline\hline
Species&Accession&Seq.length&Family&Genus \\ [0.5ex]
\hline
\textit{Chiloensis}&NC\_019601&155603 bp &\textit{Rosaceae} &\textit{Fragaria}\\
\textit{Bracteata}&NC\_018766&129788 bp &\textit{Rosaceae} &\textit{Fragaria}\\ 
\textit{Vesca}&NC\_015206&155691 bp &\textit{Rosaceae} &\textit{Fragaria}\\
\textit{Virginiana}&NC\_019602&155621 bp &\textit{Rosaceae} &\textit{Fragaria}\\
\textit{Kansuensis}&NC\_023956&157736 bp &\textit{Rosaceae} &\textit{Prunus}\\
\textit{Persica}&NC\_014697&157790 bp &\textit{Rosaceae} &\textit{Prunus}\\
\textit{Pyrifolia} &NC\_015996 &159922 bp &\textit{Rosaceae} &\textit{Pyrus} \\
\textit{Rupicola}&NC\_016921&156612 bp &\textit{Rosaceae} &\textit{Pentactina}\\
\textit{Indica}&NC\_008359 &158484 bp &\textit{Moraceae}&\textit{Morus}\\
\textit{Mollissima}&NC\_014674 &160799 bp &\textit{Fagaceae} &\textit{Castanea}\\
\hline
\end{tabular}}
\end{table}

As some branches are not well supported, we can wonder whether a few genes can be incriminated in this lack of support, for a large variety of reasons encompassing homoplasy, stochastic errors, undetected paralogy, incomplete lineage sorting, horizontal gene transfers, or even hybridization. If so, we face the optimization problem presented previously: \emph{find the most supported tree using the largest subset of core genes}. 

Genes of the core genome are now supposed to be lexicographically ordered. Each subset $S$ of the core genome is thus associated with a unique binary word $W$ of length $n$: for each $i$, $1\le i \le n$, $W_i$ is 1 if the $i$-th core gene is in $S$ and 0 otherwise. Any  $n$-length  binary word $W$ can be associated with its percentage $p$ of 1's and the lowest bootstrap $b$ of the phylogenetic tree we obtain when considering the subset of genes associated to $W$. Each word $W$ is thus associated with a fitness score value $\mathcal{F}=\frac{b+p}{2}$.

In the BPSO context the search space is then $\{0,1\}^N$, where $N=82$ in \emph{Rosales}. Each node of this $N$-cube is associated with the set of following data: its subset of core genes, the deduced phylogenetic tree, its lowest bootstrap $b$ and the percentage $p$ of considered core genes, and, finally, the score $\frac{b+p}{2}$. Notice that two close nodes of the $N$-cube have two close percentages of core genes.  We thus have to construct two phylogenies based on close sequences, leading with a high probability to the same topology with close bootstraps. In other words, the score remains essentially unchanged when moving from a node to one of its neighbors. It allows to find optimal solutions using approaches like BPSO.

During swarm initialization, the $L$ particles (set to 10 in our experiments) of a swarm are randomly distributed among all the vertices (binary words) of the $N$-cube that have a large percentage of 1's. The objective is then to move these particles in the cube so that they will converge to an optimal node. 

At each iteration, the particle velocity is updated by taking into account its own best position and the best one considering the whole particle swarm (both identified according to the fitness value). It is influenced by constant weight factors as expressed in Equation~\eqref{eq:2}. In this one, we have set $\phi_1=c_1 \cdot r_1$ and $\phi_2=c_2 \cdot r_2$ where $c_1=1$ and $c_2=1$, while $r_1$, $r_2$ are random numbers belonging to [0.1,0.5], and $w$ is the inertia weight that is computed based on the following formula:

\begin{eqnarray}\label{eq:inertia}
w=w_{max}-\frac{w_{max}-w_{min}}{I_{\text{max}}} \times I'_{\text{cur}}
\end{eqnarray}

\noindent where $I_{\text{max}}$ represents the maximum number of iterations (or time step) and $I'_{\text{cur}}$ is the current iteration. This equation determines the contribution rate of a particle's previous velocity and is determined as in~\cite{premalatha2009hybrid}.

To increase the number of included components in a particle, we reduce the interval of Equation~\eqref{eq:2} to [0.1, 0.5]. For instance, if the velocity $V_{ij}$ of an element is equal to 0.51 and $r_{ij}=0.83$, then $Sig(0.51)=0.62$. So $r_{ij}>Sig(V_{ij})$ and this leads to 0 in the vector element~$j$ of the particle~$i$. By minimizing the interval, we increase the probability of having $r_{ij}<Sig(V_{ij})$ and consequently the number of 1s, which means more included elements in the particle (a larger number of core genes).

Note that a large inertia weight facilitates a global search, while a small inertia weight tends more to a local investigation~\cite{SwarmIntell}. In other words, a larger value of $w$ facilitates a complete exploration, whereas small values promote exploitation of areas. This is why Eberhart and Shi \cite{eberhart2001particle} suggested to decrease $w$ over time, typically from 0.9 to 0.4,  thereby gradually changing from exploration to exploitation.
Finally, each particle position is updated according to 
Equation~\eqref{eq:3}.

\subsection{Distributed BPSO with MPI}\label{MPI} 

\begin{figure}[tb]
\centering
\includegraphics[scale=.7]{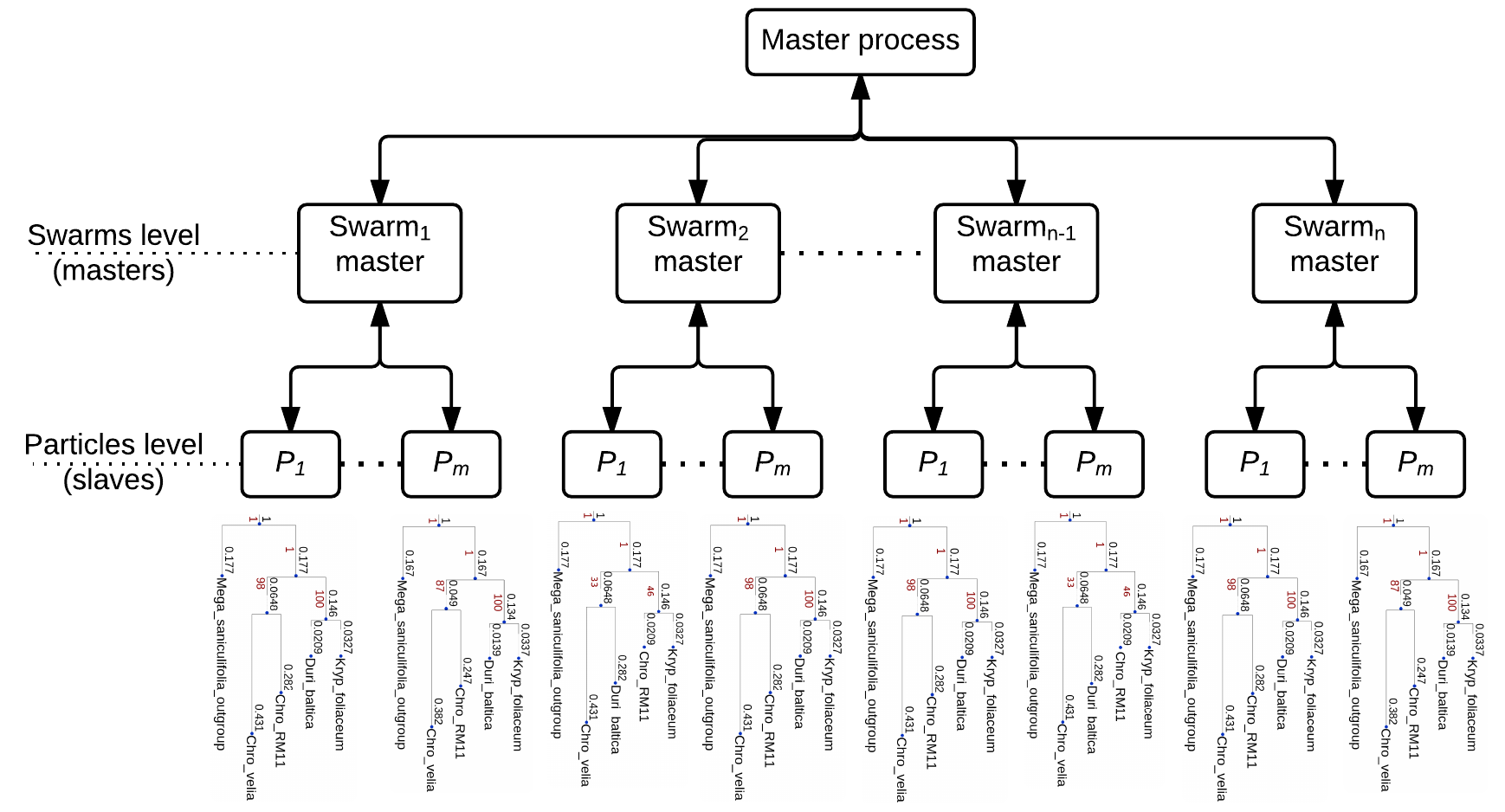}
\caption{The distributed structure of BPSO algorithm.}\label{fig:pso_parallel}
\end{figure}

Traditional PSO algorithms are time consuming in sequential mode. The distributed version shown in Figure~\ref{fig:pso_parallel}  has thus been proposed to minimize the execution time as much as possible. The general idea of the proposed algorithm is simple: 
a processor core is employed for each particle in order to compute its fitness value, while a last core called the master centralizes the obtained results. 
In other words, if we have a swarm of ten particles, we use ten cores as workers and one core as master~(or supervisor). 

More precisely, the master initializes the particles of the swarm and distributes them to the workers.
When one worker finishes its job, it sends a ``terminate'' signal with the fitness value to the master. This latter waits until all the workers have finished their jobs. Then, it determines the position of the particle that has the best fitness value as the global best position and sends this information to the workers that update their respective particle velocity and position. This mechanism is repeated until a particle achieves a fitness value larger than or equal to 95 with a large set of included genes. In the following, two distributed versions of the BPSO described previously are considered: in version~I the equation used to update the velocity is slightly changed as shown below, and in version~II we use the equations of Section~\ref{Sec:PSO}.   



\subsubsection{Distributed BPSO Algorithm: Version I} \label{subsec:version1}


In this version Equation~(\ref{eq:2}), which is used to update the velocity vector, is replaced by:
\begin{eqnarray}\label{eq:5}
V_i(t+1)= x \cdot [V_i(t)+C_1(P_i^{best}-X_i)+C_2(P_g^{best}-X_i)]
\end{eqnarray}
where $x$, $C_1$, and $C_2$ are weighted parameters setting the level of each three trends for the particle.
The default values of these parameters are $C_1=c_1 \cdot r_1=2.05$, $C_2=c_2 \cdot r_2=2.05$, while
$x$ which represents the constriction coefficient is computed according to formula~\cite{sedighizadeh2009particle,clerc1999swarm}:



\begin{eqnarray}\label{eq:constriction}
x=\frac{2 \times k}{|2-C-(\sqrt{C \times (C-4)})|},
\end{eqnarray}

\noindent where $k$ is a random value between [0,1] and $C=C_1+C_2$, where $C\geq 4$. According to Clerc~\cite{clerc1999swarm}, using a constriction coefficient results in particle convergence over time. 

\subsubsection{Distributed BPSO Algorithm: Version II} \label{subsec:version2}

This version is a distributed approach of the sequential PSO algorithm presented previously in Section~\ref{Sec:PSO}. 

\section{Phylogenetic Prediction}\label{sec:exp}

\subsection {Genetic algorithm evaluation on a large group of plant species}
\label{GAsimuls}

The proposed pipeline has been tested with the genetic algorithm on various sets of close plant species. 50 subgroups, including on average from 12 to 15 chloroplasts species, encompassing 356 plant species, and already presented in this document (\textit{c.f.} Table~\ref{tab:families}) have been used with our formerly published genetic algorithm. 
Obtained results with details are contained too in Table~\ref{tab:families}. Column \textit{Occ} represents the amount of generated phylogenetic trees from the corresponding search space for each group. The column \textit{\emph{$c$}} represents the number of core genes included within each group. The \textit{\# taxa} column is the amount of species corresponding to the considered group. \textit{\emph{$b$}} is the lowest value from bootstrap analysis. The \textit{Terminus} column contains the termination stage for each subgroup, namely: the systematic~(1), random~(2), or optimization~(3) stage using genetic algorithm and/or Lasso test. 
These stages, which have been proposed in~\cite{genetic2015}, correspond to the systematic deletion of 0 or 1 gene ($N+1$ computations for $N$ core genes), random suppression of core genes (ranging from 2 to 5 genes), and the so-called genetic algorithm on binary word populations improved by the use of a statistical test. 
Finally, the \textit{Likelihood} column stores the likelihood value of the best phylogenetic tree (\textit{i.e.}, according to the lowest bootstrap value $b$). A large occurrence value in this table means that the associated $p$-value and/or subgroup has its computation terminated in either penultimate or last pipeline stage. An occurrence of 31 is frequent due to the fact that 32 MPI threads (one master plus 31 slaves) have been launched on our supercomputing facility.

Notice that the groups in Table~\ref{tab:families} can be divided in four parts:
\begin{itemize}
\item Groups of species stopped in systematic stage with weak bootstrap values. This is due to the fact that an upper time limit has been set for each group and/or subgroups, while each computed tree in these remarkable groups needed a lot of times for computations.
\item Subgroups terminated during systematic stage with desired bootstrap value.
\item Groups or subgroups terminated in random stage with desired bootstrap value.
\item Finally, groups or subgroups terminated with optimization stages.
\end{itemize}
A majority of subgroups has its phylogeny satisfactorily resolved, as can be seen on all obtained trees which can be downloadable at \url{http://meso.univ-fcomte.fr/peg/phylo}.
However, some problematic subgroups still remain to be investigated, which explains why the distributed BPSO is considered in the next section.

\subsection{First experiments on \textit{Rosales} order}

\begin{figure}[tb]
\CenterFloatBoxes
\begin{floatrow}
\ffigbox
  {\includegraphics[scale=0.35]{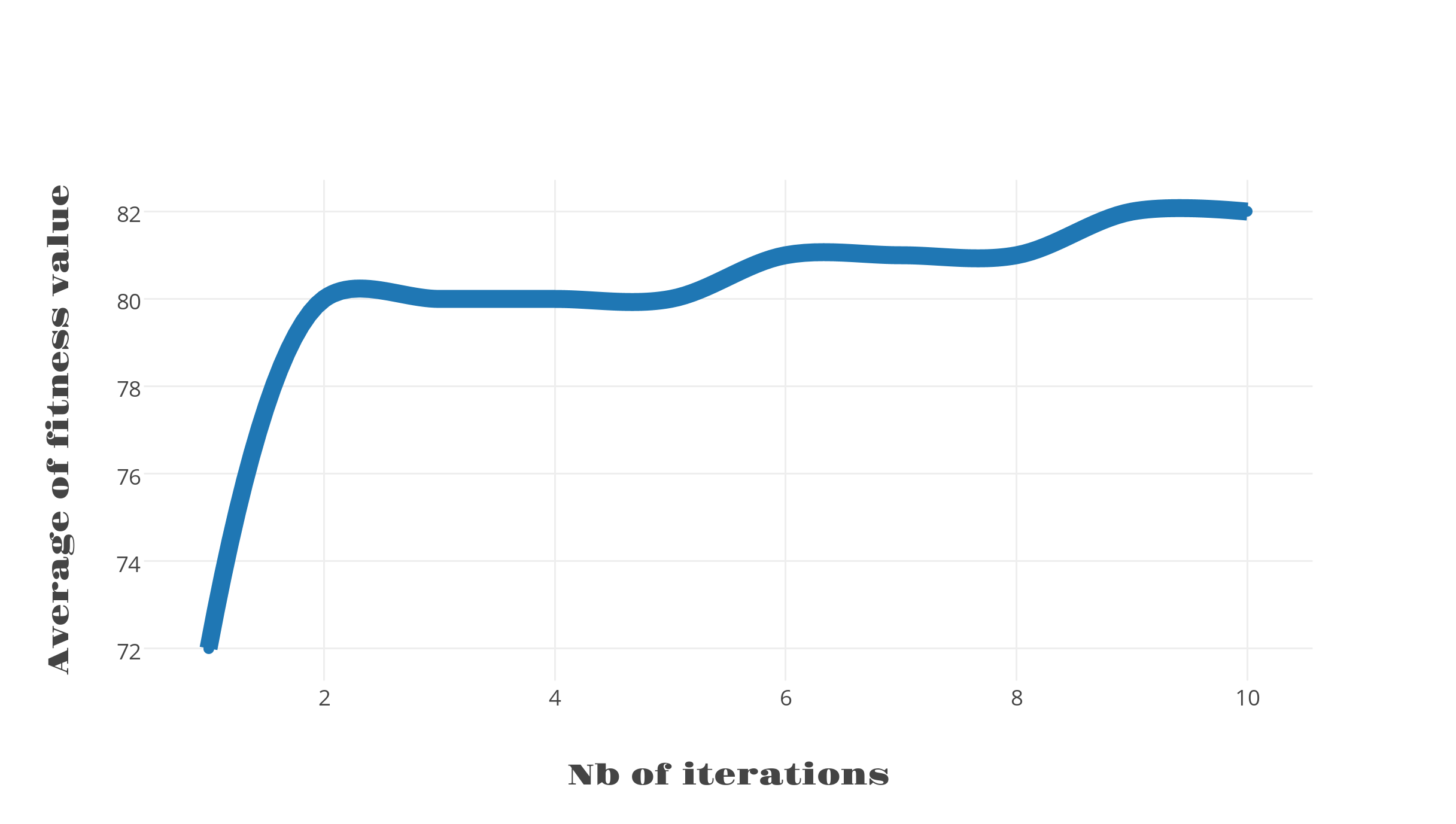}}
  {\caption{Average fitness of \textit{Rosales} order}\label{overflow first version}}
\killfloatstyle
\ttabbox
{\hspace*{1cm}\scalebox{0.8}{%
  \begin{tabular}{c | c| c | c  } 
&Removed & &  \\
Swarm & genes & $\mathcal{F}$ & $b$ \\ [0.5ex]
\hline 
1 &4 &75.5&73\\
2 &6  &75.5&76\\
3 &20 &75  &88 \\
4 &52 &59.5&89 \\
5 &3  &75.5&72 \\
6 &19 &77.5&92 \\
7 &47 &63.5&92 \\
8 &9 &73.5&74 \\
9 &10 &72.5 &73 \\
10&13 &76.5 &84 \\
\end{tabular}}
  }
{\caption{Best tree in each swarm\label{tab:table2}}}
\end{floatrow}
\end{figure}

In a first collection of experiments, we have implemented the proposed BPSO algorithm on a supercomputing facility. Investigated species are the ones listed in Table~\ref{tab:species}.
10 swarms having a variable number of particles have been launched 10 times, with $c_1=1, c_2=1$, and $w$ 
 linearly decreasing from 0.9 to 0.4. 
Obtained results are summarized in Table~\ref{tab:table2} that contains, for each 10 runs of each 10 swarms: the number of removed genes and the minimum bootstrap of the best tree. Remark that some bootstraps are not so far from the intended ones (larger than 95), whereas the number of removed genes are in average larger than what is desired. 

\begin{figure}[tb]
\centering
\subfigure[$Topology_0$]{\includegraphics[scale=0.04]{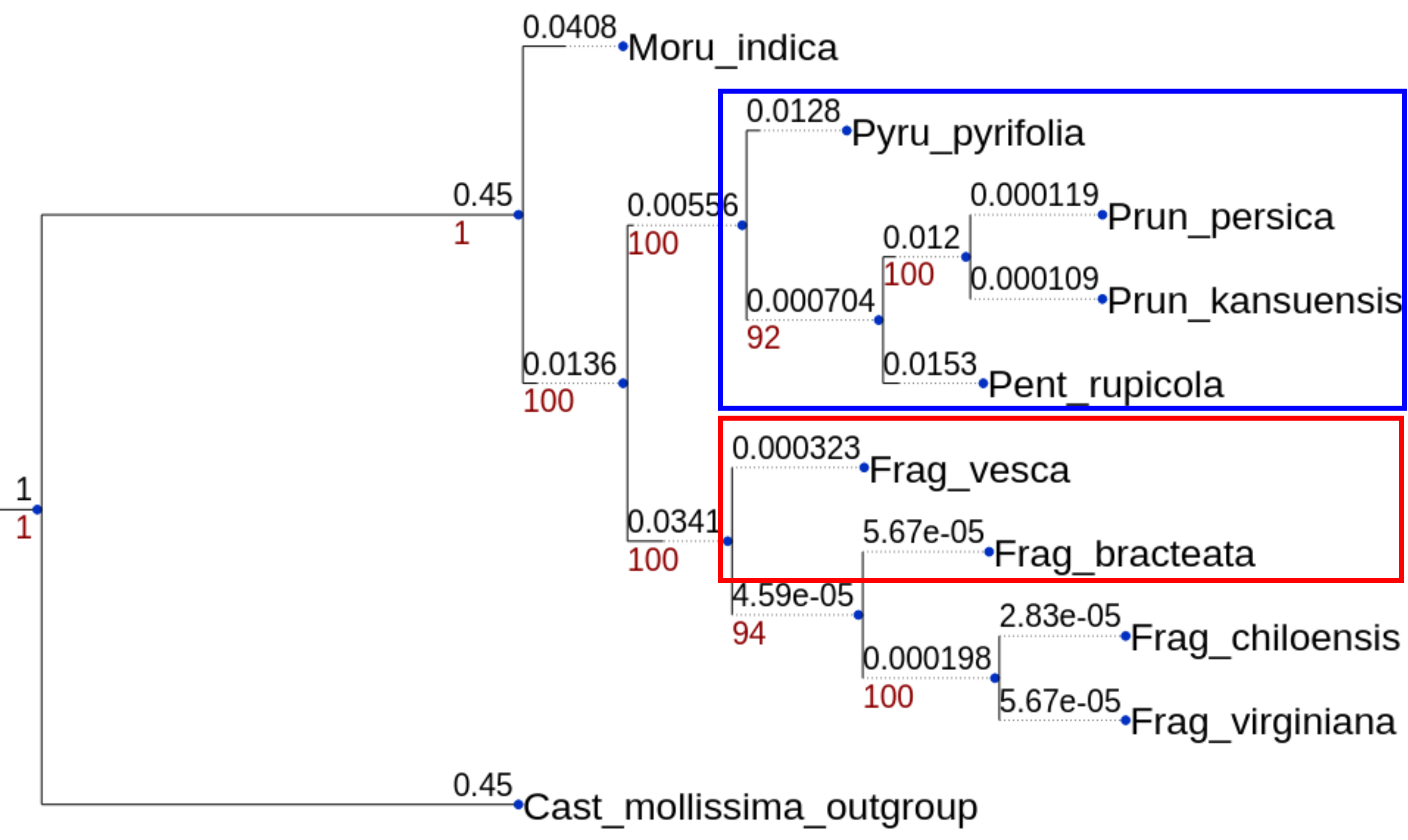}\label{Topology0}}
\subfigure[$Topology_4$]{\includegraphics[scale=0.04]{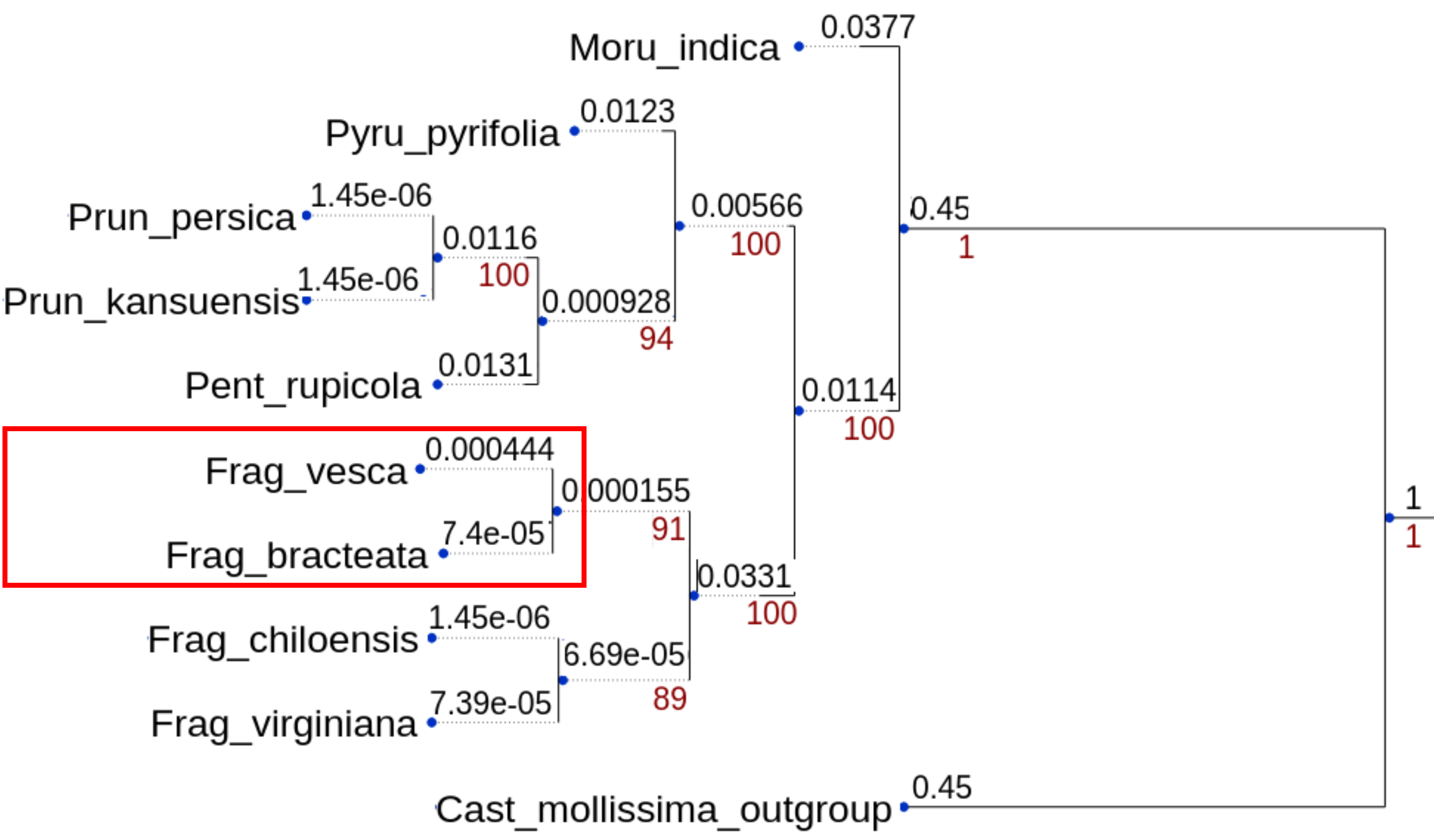}\label{Topology4}}\\
\subfigure[$Topology_2$]{\includegraphics[scale=0.04]{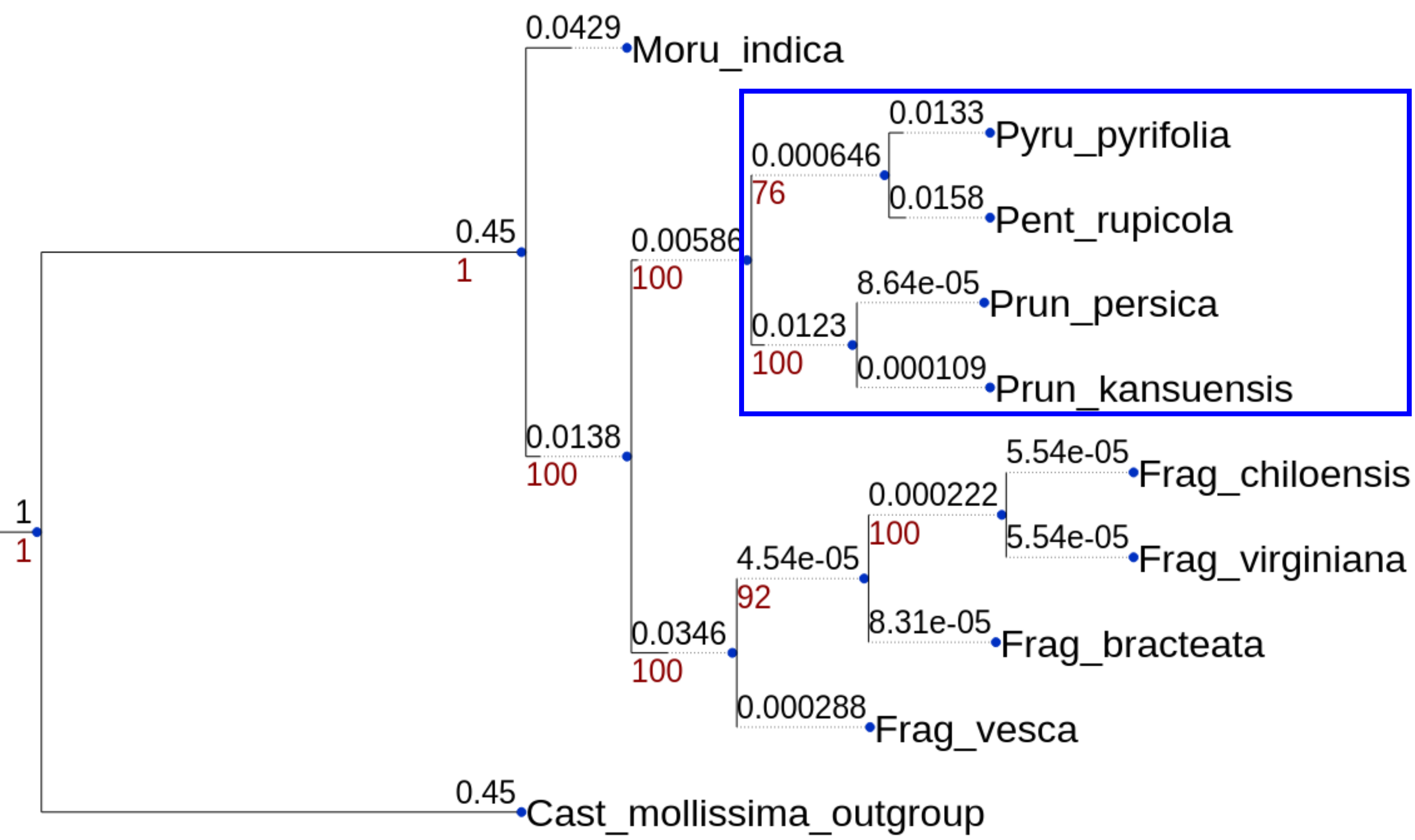}\label{Topology2}}
\caption{The best obtained topologies for~\emph{Rosales} order}
\label{fig:topo}
\end{figure}

Seven topologies have been obtained after either convergence or $maxIter$ iterations. Only 3 of them have occurred a representative number of times, namely the Topologies 0, 2, and 4, which are depicted in Figure~\ref{fig:topo} (see details in Table~\ref{tab:table3}). 

\begin{table}[!ht]
\caption{Best topologies obtained from the generated trees, $b$ is the lowest bootstrap of the best tree having this topology,= $p$ is the number of considered genes to obtain this tree.}\label{tab:table3}
\centering
\scalebox{0.8}{
\begin{tabular}{c|c | c | c | c |c} 
Topology & Swarms & $b$ & $p$ & $\mathcal{F}$ & Occurrences\\[0.5ex]
\hline 
0 & 1, 2, 3, 4, 5, 6, 7, 8, 9, 10     &92  &63 &77.5  &568 \\
1 & 1, 2, 3, 4, 5, 6, 10              &63  &45 &54    &11  \\
2 & 1, 2, 3, 4, 5, 6, 7, 8, 9, 10     &76  &67 &71.5  &55  \\  
3 & 8, 1, 2, 3, 4                     &56  &41 &48.5  &5   \\
4 & 1, 2, 3, 4, 5, 6, 7, 8, 9, 10     &89  &30 &59.5  &65  \\ 
5 & 1, 3, 4, 5, 6, 9                  &71  &33 &52    &9   \\
6 & 5, 6                              &25  &45 &35    &2   \\
\end{tabular}}
\end{table}

These three topologies are almost well supported, except in a few branches. We can notice that the differences in these topologies are based on the sister relationship of two species named \emph{Fragaria vesca} and \emph{Fragaria bracteata}, and of the relation between \emph{Pentactina rupicola} and \emph{Pyrus pyrifolia}.
Due to its larger score and number of occurrences, we tend to select Topology~0 as the best representative of the \emph{Rosale} phylogeny. 

To further validate this choice, CONSEL~\cite{shimodaira2001consel} software has been used on per site likelihoods of each best tree obtained using the RAxML~\cite{Stamatakis21012014}. The CONSEL computes the $p$-values of various well-known statistical tests, like the so-called approximately unbiased (au), Kishino-Hasegawa (kh), Shimodaira-Hasegawa (sh), and Weighted Shimodaira-Hasegawa (wsh) tests. Obtained results are provided in Table~\ref{tab:consel1}, they confirm the selection of Topology~0 as the tree reflecting the best the \emph{Rosales} phylogeny. 
\begin{table}[tb]
\footnotesize
\centering
\caption{The CONSEL results regarding best trees\label{tab:consel1}}
\begin{tabular}{c|c|c|c|c||c|c|c|c|c|c}
Rank & item &   obs &     au &     np &     bp &     pp &     kh &     sh &    wkh &    wsh \\ \hline
1 & 0 & -1.4 & 0.774 & 0.436 & 0.433 & 0.768 & 0.728 & 0.89 & 0.672 & 0.907 \\ 
2 & 4 & 1.4 & 0.267 & 0.255 & 0.249 & 0.194 & 0.272 & 0.525 & 0.272 & 0.439 \\ 
3 & 2 & 3 & 0.364 & 0.312 & 0.317 & 0.037 & 0.328 & 0.389 & 0.328 & 0.383 \\ 
\end{tabular}
\end{table}

After having verified that BPSO can be used to resolve phylogenetic issues thanks to the \textit{Rosales} order, we now intend to deeply compare the genetic algorithm versus the swarm particle optimization. In order to do so, a large collection of group of plant species have been selected, on which we have successively launched the genetic algorithm and the BPSO one in distributed mode.

\subsection{Comparison between distributed version of GA and the two distributed versions of BPSO}\label{sec:GA}

12 groups of plant genomes have been extracted from the 49 ones used in the GA evaluation. More precisely, seven ``difficult'' groups have been selected from those that have reached the third stage in genetic algorithm method (no resolution of phylogeny during systematic and random modes). Conversely, five ``easy'' groups have been added in the pool of experiments, for the sake of comparison: in these groups, the phylogeny has been resolved during the systematic mode. They have been applied on our two swarm versions, and results have been compared to the genetic algorithm ones. We have successively tested 10 and 15 particles (with each of the two algorithms), on the supercomputer facilities.

Comparisons are provided in Tables~\ref{tab:3} and~\ref{tab:4}. In these tables, 
\textit{Topo.} column stands for the number of topologies, 
\textit{NbTrees} is the total number of obtained trees using 10 swarms,
$b$ is the minimum bootstrap value of selected $w$, $100-p$ is the number of missing genes in $w$ 
and \textit{Occ.} is the number of occurrences of the best obtained topology from 10 swarms.
As can be seen in these tables, 
the two versions of BPSO did not provide the same kind of results:
\begin{itemize}
\item In the case of \textit{Chlorophyta}, \textit{Pinus}, and \textit{Bambusoideae}, the second version of the BPSO has outperformed the first one, as the minimum bootstrap $b$ of the best tree is finally larger for at least one swarm.
\item In the \textit{Ericales} case, the first version has produced the best result.
\end{itemize}

\begin{figure}[tb]
\CenterFloatBoxes
\begin{floatrow}
\ttabbox
{    \scalebox{0.48}{%
    \begin{tabular}{cccccccccc}
        \hline\hline
        Group  &Topo. &NbTrees &$b$    &$|c|$   &$100-p'$    &Occ.  &Swarms &Particles\\
        \hline
        \textit{Pinus} &3  &508  &98 &79     &32   &462	 &{1,2,3,4,5,6,7,8,9,10}   &10\\
        \textit{Pinus} &3  &530  &94 &79     &11   &129	 &{1,2,3,4,5,6,7,8,9,10}   &15\\
        \textit{Picea} &1  &100  &100	&85	&42	   &100	 &{1,2,3,4,5,6,7,8,9,10}   &10\\
        \textit{Picea} &1 &428 &100 &85 &13 &428 &{1,2,3,4,5,6,7,8,9,10} &15\\ 
        \textit{Magnoliidae} &3  &750 &100  &79 &20 &613&{1,2,3,4,5,6,7,8,9,10} &10\\
        \textit{Magnoliidae} &3  &845 &100  &79 &19 &707&{1,2,3,4,5,6,7,8,9,10} &15\\
        \textit{Ericales} &30  &344  &53	&84 &26 &185  &{1,2,3,4,5,6,7,8,9,10} &10\\
        \textit{Ericales} &34  &555  &54	&84 &5	&363  &{1,2,3,4,5,6,7,8,9,10} &15\\
        \textit{Bambusoideae} &8 &496   &72	&94 &37	&456 &{1,2,3,4,5,6,7,8,9,10}   &10\\
        \textit{Bambusoideae} &11  &694 &69	&94	&18 &621 &{1,2,3,4,5,6,7,8,9,10}   &15\\
        \textit{Eucalyptus}  &16 &828 &86 &83 &7 &632 &{1,2,3,4,5,6,7,8,9,10} &10\\
        \textit{Eucalyptus}  &20 &1073 &86 &80 &4 &845 &{1,2,3,4,5,6,7,8,9,10} &15\\
        \textit{Malpighiales}&34 &327  &65	&78	&35	&233  &{1,2,3,4,5,6,7,8,9,10} &10 \\
        \textit{Malpighiales} &38  &483 &69	&78	&40	&326  &{1,2,3,4,5,6,7,8,9,10} &15\\
        \textit{Chlorophyta}&25 &191 &70	&24	&11 &109  &{1,2,3,4,5,6,7,8,9,10} &10\\
        \textit{Chlorophyta}&29 &94 &68 &24  &11 &1 &{1,2,3,4,5,6,7,8,9,10} &15\\
        \textit{Euglenozoa}  &3  &450 &100 &26 &7  &292 &{1,2,3,4,5,6,7,8,9,10} &10\\
        \textit{Euglenozoa}  &3  &520 &100 &26 &4  &491 &{1,2,3,4,5,6,7,8,9,10} &15\\
        \textit{Ehrhartoideae}&2 &23 &100 &81 &0  &23 &{1,2,3,4,5,6,7,8,9,10} &10\\
        \textit{Ehrhartoideae}&3 &455 &100 &81 &0 &451 &{1,2,3,4,5,6,7,8,9,10} &15\\
        \textit{Trebouxiophyceae}&3 &409 &100 &41 &2 &405 &{1,2,3,4,5,6,7,8,9,10} &10\\
        \textit{Trebouxiophyceae}&3 &415 &100 &41 &8 &354 &{1,2,3,4,5,6,7,8,9,10} &15\\
        \textit{Poeae}&1 &971 &100 &80 &9 &971 &{1,2,3,4,5,6,7,8,9,10}&10\\
        \textit{Poeae}&1 &1399 &100 &80 &20 &1399 &{1,2,3,4,5,6,7,8,9,10}&15\\
        \hline 
        \end{tabular}}}
        {\caption{Groups from BPSO Version~1.\label{tab:3}}}
\killfloatstyle
\ttabbox
 {\hspace*{1cm}\scalebox{0.48}{%
        \begin{tabular}{cccccccccc}
        \hline\hline
        Group  &Topo. &NbTrees &$b$    &$|c|$   &$100-p'$    &Occ.  &Swarms &Particles\\
        \hline
        \textit{Pinus} &3  &615 &98  &79 &14 &275 &{1,2,3,4,5,6,7,8,9,10}&10\\
        \textit{Pinus} &3  &628 &100  &79 &12 &558 &{1,2,3,4,5,6,7,8,9,10}&15\\
        \textit{Picea} &1  &635 &100 &85 &14  &635 &{1,2,3,4,5,6,7,8,9,10} &10\\
        \textit{Picea} &1  &821 &100 &85 &15&821  &{1,2,3,4,5,6,7,8,9,10} &15\\
        \textit{Magnoliidae} &3  &494 &100  &79 &16 &73&{1,2,3,4,5,6,7,8,9,10}&10\\
        \textit{Magnoliidae} &3  &535 &100  &79 &42 &384&{1,2,3,4,5,6,7,8,9,10}&10\\
        \textit{Bambusoideae}&6  &952 &84  &81 &23 &94&{1,2,3,4,5,6,7,8,9,10}&10\\
        \textit{Bambusoideae}&9  &1450 &82  &81 &18 &113&{1,2,3,4,5,6,7,8,9,10}&15\\
        \textit{Eucalyptus}  &17 &972 &88  &80 &18 &618 &{1,2,3,4,5,6,7,8,9,10}&10\\
        \textit{Eucalyptus}  &23 &1439 &92 &80 &10 &843 &{1,2,3,4,5,6,7,8,9,10}&15\\
        \textit{Chlorophyta} &25 &529 &71  &24 &6  &397&{1,2,3,4,5,6,7,8,9,10}&10\\
        \textit{Chlorophyta} &46 &1500 &82  &24 &11  &397&{1,2,3,4,5,6,7,8,9,10}&10\\
        \textit{Ericales}    &30 &97  &51  &84 &11 &56&{1,2,3,4,5,6,7,8,9,10}&10\\
        \textit{Ericales}    &34 &1257&52  &84 &7  &800&{1,2,3,4,5,6,7,8,9,10}&15\\
        \textit{Malpighiales}&35  &725 &72 &79 &25 &445 &{1,2,3,4,5,6,7,8,9,10}&10 \\
        \textit{Malpighiales}&86 &1464 &84 &79 &45 &359 &{1,2,3,4,5,6,7,8,9,10}&15\\
        \textit{Euglenozoa}  &3  &197 &100 &26 &1  &165 &{1,2,3,4,5,6,7,8,9,10}&10\\
        \textit{Euglenozoa}  &3  &450 &100 &26 &10  &393 &{1,2,3,4,5,6,7,8,9,10}&15\\
        \textit{Ehrhartoideae}&1 &24 &100 &81 &10  &24 &{1,2,3,4,5,6,7,8,9,10}&10\\
        \textit{Ehrhartoideae}&1 &20 &100 &81 &9  &20 &{1,2,3,4,5,6,7,8,9,10}&15\\
        \textit{Trebouxiophyceae}&3 &319 &100 &41 &1 &313 &{1,2,3,4,5,6,7,8,9,10}&10\\
        \textit{Trebouxiophyceae}&3 &818 &100 &41 &2 &81 &{1,2,3,4,5,6,7,8,9,10}&15\\
        \textit{Poeae}&1 &991 &100 &80 &22 &991 &{1,2,3,4,5,6,7,8,9,10}&15\\
        \textit{Poeae}&1 &1490 &100 &80 &26 &1490 &{1,2,3,4,5,6,7,8,9,10}&15\\
        \hline 
        \end{tabular}}}
        {\caption{Groups from BPSO Version~2.\label{tab:4}}}
\end{floatrow}
\end{figure}

\begin{table}[tb]
\tiny
\caption{PSO vs GA.}\label{tab:gavspso}
\begin{center}
\begin{tabular}{l|c|c|c|c|c}
\hline
\multicolumn{1}{c|}{} & \multicolumn{ 2}{c|}{\textbf{PSO Ver.I}} & \multicolumn{ 2}{c|}{\textbf{PSO Ver.II}} & \\ \hline
\textbf{Group} & \textbf{10} & \textbf{15} & \textbf{10} & \textbf{15} & \textbf{GA} \\ \hline
\textit{Ericales} & 53 & 54 & 51 & 52 & 67 \\ 
\textit{Bambusoideae} & 72 & 69 & 84 & 82 & 80 \\ 
\textit{Pinus} & 98 & 94 & 98 & 100 & 80 \\ 
\textit{Chlorophyta} & 70 & 68 & 71 & 82 & 81 \\ 
\textit{Eucalyptus} & 86 & 86 & 88 & 92 & 90 \\ 
\textit{Malpighiales} & 65 & 69 & 72 & 84 & 96 \\ 
\textit{Magnoliidae} & 100 & 100 & 100 & 100 & 98 \\ 
\textit{Ehrhartoideae} & 100 & 100 & 100 & 100 & 100 \\ 
\textit{Euglenozoa} & 100 & 100 & 100 & 100 & 100 \\ 
\textit{Picea} & 94 & 100 & 100 & 100 & 100 \\ 
\textit{Poeae} & 80 & 80 & 100 & 100 & 100 \\ 
\textit{Trebouxiophyceae} & 100 & 100 & 100 & 100 & 100 \\ \hline
\end{tabular}
\end{center}
\end{table}

\begin{figure}[tb]
\centering
\subfigure[PSO with 15 particles vs. GA]{\includegraphics[width=60mm]{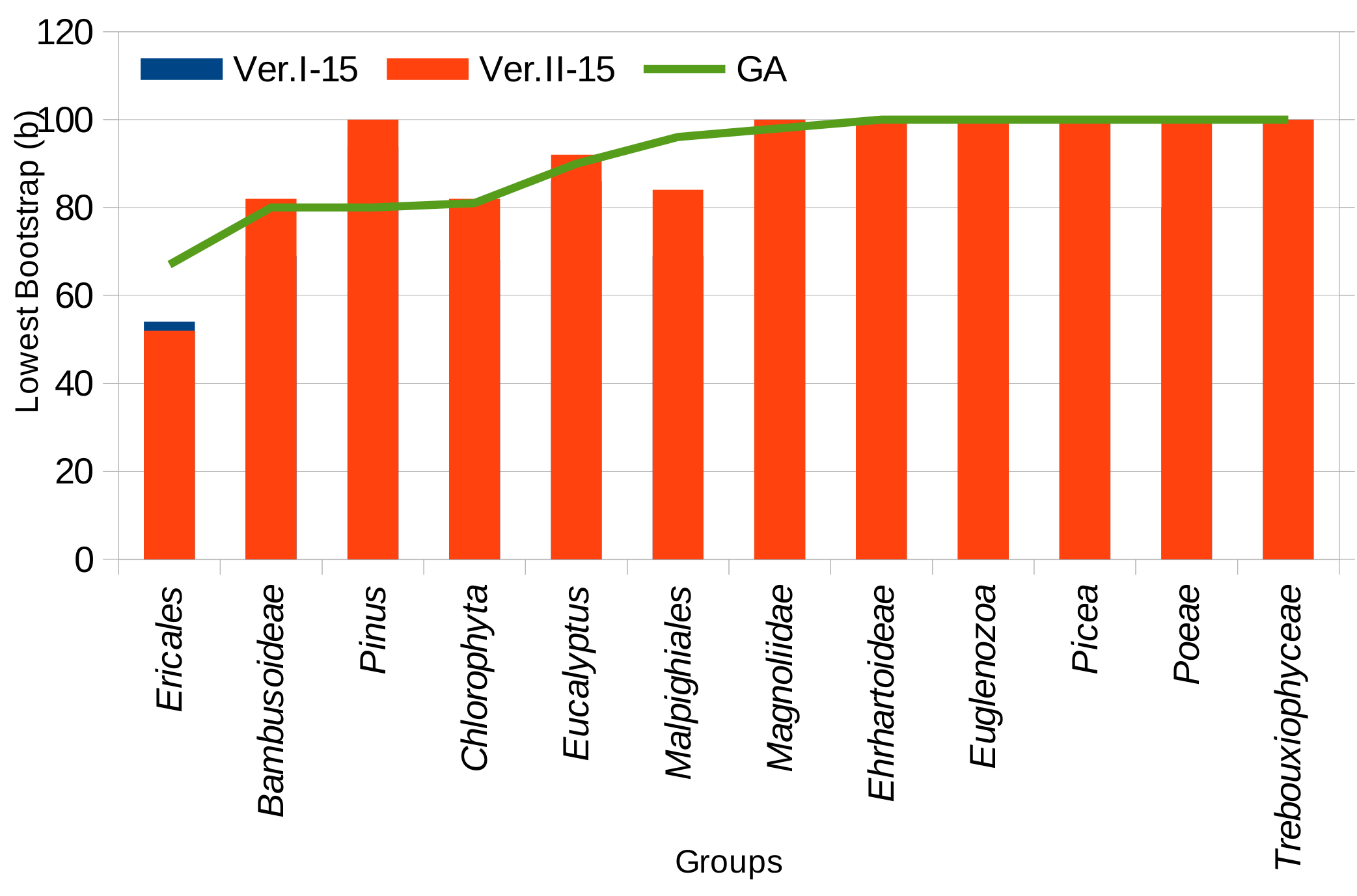}\label{f1}}
\subfigure[PSO with 10 particles vs. GA]{\includegraphics[width=60mm]{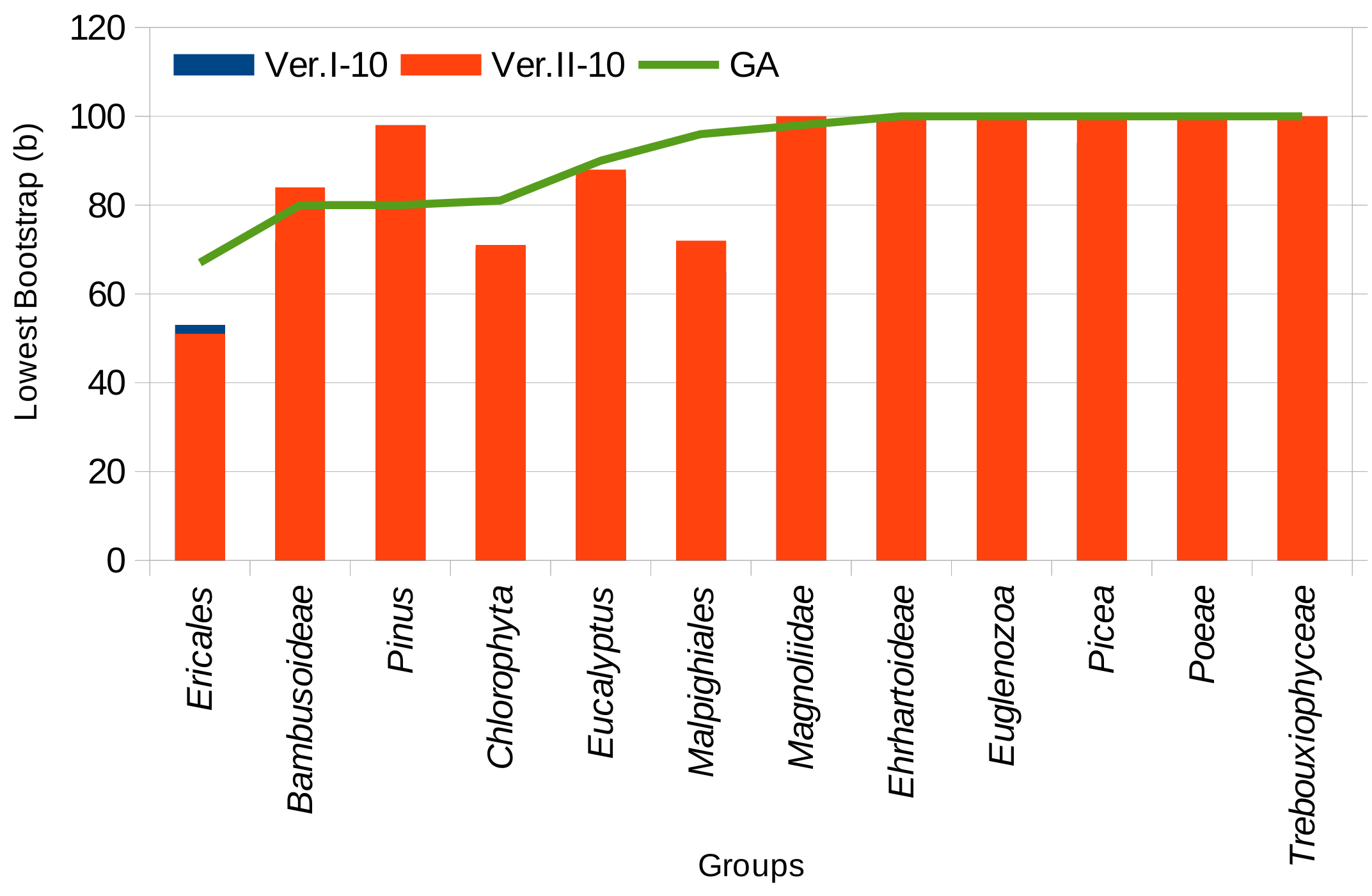}\label{f2}}
\caption{\textbf{PSO with 10 and 15 particles vs. GA.}}
\label{fig:psoandGA}
\end{figure}

We can also remark that \textit{Malpighiales} has better $b$ in GA than the two versions of PSO. For easy to solve subgroups, \textit{Pinus} data set has got maximum bootstrap~$b$ larger than what has been obtained using the genetic algorithm, while \textit{Picea} and \textit{Trebouxiophyceae} have got the same values of $b$ than in genetic algorithm. More comparison results between GA and both versions of PSOs are provided in Figure~\ref{fig:psoandGA}.

According to this figure, we can conclude that the two approaches lead to quite equivalent bootstrap values in most data sets, while on particular subgroups obtained results are complementary. In particular,  PSO often produces better bootstraps that GA (see \textit{Magnoliidae} or on~\textit{Bambusoideae}), but with a larger number of removed genes. Finally, using 15 particles instead of 10 does not improve so much the obtained results (see Figure~\ref{fig:psoandGA} and Table~\ref{tab:gavspso}).


\section{Conclusion} 

This article has presented an original method to produce a well supported 
and large-scale phylogenetic tree of chloroplast species where various optimization algorithms are applied to highlight the relationships among given gene sequences.

More precisely, this method first discovers and removes blurring genes in the set of core genes by applying a bootstrap analysis for each tree produced from a subset of core genes. 
It then continues with integrating a discrete PSO method to provide the largest subset of sequences. Two distributed versions of this PSO-based optimization step
have been developed in order to reduce the computation time and memory used. 
Finally, a per site analysis by the CONSEL is applied:
a dedicated topological process analyses all the output trees and 
might use a per site analysis in order to extract the most 
relevant ones. Our proposed pipeline has been applied to various families of plant species. More than 65\% of phylogenetic trees produced by this pipeline have presented bootstrap values larger than 95.




\section*{Acknowledgements}
All computations have been performed on the \emph{M\'esocentre de calculs} supercomputer facilities of the University of Franche-Comt\'e.

\bibliographystyle{plain}
\bibliography{biblio}

\end{document}